\documentclass[10pt,journal]{IEEEtran}
\IEEEoverridecommandlockouts
\usepackage{cite}
\usepackage{amsmath,amssymb,amsfonts}

\usepackage{algorithm}
\usepackage{algorithmicx}
\usepackage{algpseudocode}

\usepackage{graphicx}
\usepackage{textcomp}
\usepackage{xcolor}
\usepackage{subcaption}

\usepackage[hidelinks]{hyperref}

\usepackage{color,xcolor}
\usepackage{epsfig}
\usepackage{graphicx}

\usepackage{adjustbox}
\usepackage{array}
\usepackage{booktabs}
\usepackage{multirow}

\usepackage{amsmath,amsfonts,amsthm,amssymb}
\usepackage{mathrsfs}
\usepackage{bm}
\usepackage{nicefrac}
\usepackage{microtype}

\usepackage{changepage}
\usepackage{extramarks}
\usepackage{fancyhdr}
\usepackage{lastpage}
\usepackage{setspace}
\usepackage{soul}
\usepackage{xspace}

\usepackage[switch]{lineno}

\newlength\myindent
\def\BibTeX{{\rm B\kern-.05em{\sc i\kern-.025em b}\kern-.08em
    T\kern-.1667em\lower.7ex\hbox{E}\kern-.125emX}}
    
\begin{document}



\title{Self-supervised On-device Federated Learning from Unlabeled Streams}

\author{Jiahe Shi,
    Yawen Wu, Dewen Zeng, 
    Jun Tao,~\IEEEmembership{senior member,~IEEE,}
    Jingtong Hu,~\IEEEmembership{senior member,~IEEE,}
    \\
    Yiyu Shi,~\IEEEmembership{senior member,~IEEE,}

\thanks{Jiahe Shi and Jun Tao are with the State Key Laboratory of ASIC and System, School of Microelectronics, Fudan University, Shanghai 200433, China (e-mail: jhshi21@m.fudan.edu.cn, taojun@fudan.edu.cn).}
\thanks{Yawen Wu, Dewen Zeng and Yiyu Shi are with the Department of Computer Science and Engineering,
University of Notre Dame, Notre Dame, IN 46556 USA (e-mail: ywu37@nd.edu, dzeng2@nd.edu, yshi4@nd.edu).}
\thanks{Jingtong Hu is with the Department of
Electrical and Computer Engineering, University of Pittsburgh, Pittsburgh,
PA 15261 USA (e-mail: jthu@pitt.edu).}
}




\maketitle


\begin{abstract}
The ubiquity of edge devices has led to a growing amount of unlabeled data produced at the edge. Deep learning models deployed on edge devices are required to learn from these unlabeled data to continuously improve accuracy. Self-supervised representation learning has achieved promising performances using centralized unlabeled data. However, the increasing awareness of privacy protection limits centralizing the distributed unlabeled image data on edge devices.  
While federated learning has been widely adopted to enable distributed machine learning with privacy preservation, without a data selection method to efficiently select streaming data, the traditional federated learning framework fails to handle these huge amounts of decentralized unlabeled data with limited storage resources on edge. 
To address these challenges, we propose a Self-supervised On-device Federated learning framework with coreset selection, which we call SOFed,
to automatically select a \emph{coreset} that consists of the most representative samples into the replay buffer on each device.
It preserves data privacy as each client does not share raw data while learning good visual representations. Experiments demonstrate the effectiveness and significance of the proposed method in visual representation learning.

\end{abstract}

\begin{IEEEkeywords}
Learning Representations, On-Device Learning, Federated Contrastive Learning, Self-Supervised Learning
\end{IEEEkeywords}

\section{Introduction}

Deep learning models deployed on edge devices have made significant improvements in the last few years to extract useful visual representations in real-world tasks, such as sensors for fault diagnosis in smart manufacturing \cite{Industry2021Khalil}, and agricultural robots for fruit anomalies monitoring \cite{ICRA2022Choi}. The significance of the deep neural networks highly relies on the centralized supervised training paradigm, where the models are trained on labeled data in the server. However, a tremendous amount of unlabeled image data that cannot be centralized is produced by edge devices every day. On the one hand, the decentralized data is usually of private nature, e.g., photos taken from medical monitoring may contain patient faces. Sending the data to the server may violate data privacy regulations. On the other hand, the cost of labeling is prohibitive. Hence, it is desirable for the deep learning models deployed on edge devices to directly learn from this unlabeled decentralized data stream. Besides, compared with the traditional training method, deep learning models learn better representations for the deployed environment by leveraging local data.

To learn effective visual representations without supervision, researchers have proposed many self-supervised representation learning methods \cite{ECCV2015Unsupervised, ECCV2016Unsupervised, chen2020simclr, he2020moco, cvpr2021simsiam, NIPS2020BYOL,survey2021ssl}. Among them, contrastive learning \cite{chen2020simclr, he2020moco, cvpr2021simsiam,NIPS2020BYOL} is the state-of-the-art method. Contrastive learning methods instruct neural networks to minimize dissimilarity between representations of two similar data points or to maximize dissimilarity  between representations of two different data points. Although contrastive learning has demonstrated promising performance in learning from unlabeled data, it is conducted on centralized data in servers or distributed devices without storage restrictions.

However, it is impractical to collect a large centralized dataset from edge devices due to privacy concerns and to store all the data locally on each device due to storage overhead.
First, due to privacy regulations, it is unfeasible to directly send the data samples in the local data buffer to a high-performance server. Existing methods learning from unlabeled data \cite{chen2020simclr, he2020moco,cvpr2021simsiam, NIPS2020BYOL} cannot collaboratively leverage decentralized unlabeled image data to learn a visual representation under the restriction of limited storage resources while keeping data private.
Second, the images generated by the edge devices are usually captured by the integrated cameras in sequence and normally are non-independent-and-identically distributed (non-iid). 
Storing the constantly generated streaming data on distributed edge devices is prohibitive because of the storage and energy overhead. Therefore, it is necessary to select only a “coreset” into the small data buffer on edge device where only the most representative samples are stored.

Federated learning \cite{pmlr2017mcmahan} is a distributed training technique with privacy guaranteed. Existing works that combine contrastive learning with federated learning \cite{ICCAD2021Wu, FedCA2020zhang, ICCV2021fedu, ijcai2022wu} either have privacy leakage risks by sharing features or do not consider the restriction of limited storage volumes. To learn visual representations from unlabeled data generated by networked devices with limited on-device storage, a small data buffer can be used for each edge device to maintain the most effective data for learning. However, there are several challenges we need to address to select high-quality data into the data buffer in the federated contrastive learning (FCL) scenario.
First, privacy cannot be compromised when measuring the sample's quality and impact on the global model. The features of clients’ local data should not be exposed to other parties, including the server. Otherwise, the secrets of the participating client in the network may be revealed by others. For example, in a network of distributed industrial robots in car plants, if a robot produces a feature distinguished from others, it may reveal the manufacturing of a new vehicle model. Hence, a sample selection policy must be carefully designed to protect the privacy of clients.
Second, the selected collection of samples should be effective for the global federated learning model. The distributed unlabeled data in FCL complicate the online data selection. As a result of the lack of knowledge of other devices’ data distribution, the samples selected by different devices may overlap, hence, deviate from the distribution of training data collected by all devices.
Moreover, due to the restricted computation, storage, and communication resources of local edge devices, the selection process should introduce marginal overheads. Without an efficient data selection policy, the performance of mobile systems may degrade.

To address these challenges,
we propose a novel \textbf{S}elf-supervised \textbf{O}n-device \textbf{Fed}rated learning framework with coreset selection, SOFed,
to learn visual representations from unlabeled data streams collaboratively on edge devices. A server coordinates distributed edge devices to conduct federated contrastive learning. We use a coreset selection method based on importance scoring to maintain the most representative samples in the local buffer. A larger importance score indicates the sample has a larger impact on the model. Selecting samples with higher importance scores into the replay buffer enables the replay buffer to maintain more valuable data.
Each client conducts unsupervised representation learning on the data in the local buffer and collaboratively optimizes the global model.
After federated contrastive learning, we utilize the global model as the backbone. Then we train an additional new classifier on top of the backbone model with a limited number of labels.

In summary, the main contributions of the paper include:
\begin{itemize}
    \item \textbf{Federated on-device contrastive learning framework.} We propose an on-device FCL framework, SOFed, 
    to form a small data buffer on each edge device for collaborative self-supervised learning from real-time streaming data.
There is only a small data buffer on each edge device to eliminate unaffordable storage overhead.
\textit{This is the first work on self-supervised on-device federated learning from unlabeled streams.}

	\item \textbf{Coreset selection method in a federated contrastive learning scenario.} We propose a coreset selection method based on importance scoring to maintain the most effective data in the buffer for local training. Labels are not needed in the coreset selection and privacy is guaranteed by keeping features and raw data local.
    The computation of importance scores is efficient and induces low overhead.
	\item \textbf{Better classification accuracy and label efficiency.} 
Experiments on various public datasets show that SOFed outperforms state-of-the-art baselines with different sampling methods and self-supervised learning methods in classification accuracy and label efficiency.
\end{itemize}

\section{Background and Related Work} \label{sec:background}

\begin{figure}[!htb]
	\centering
	\includegraphics[width=1.0\linewidth]{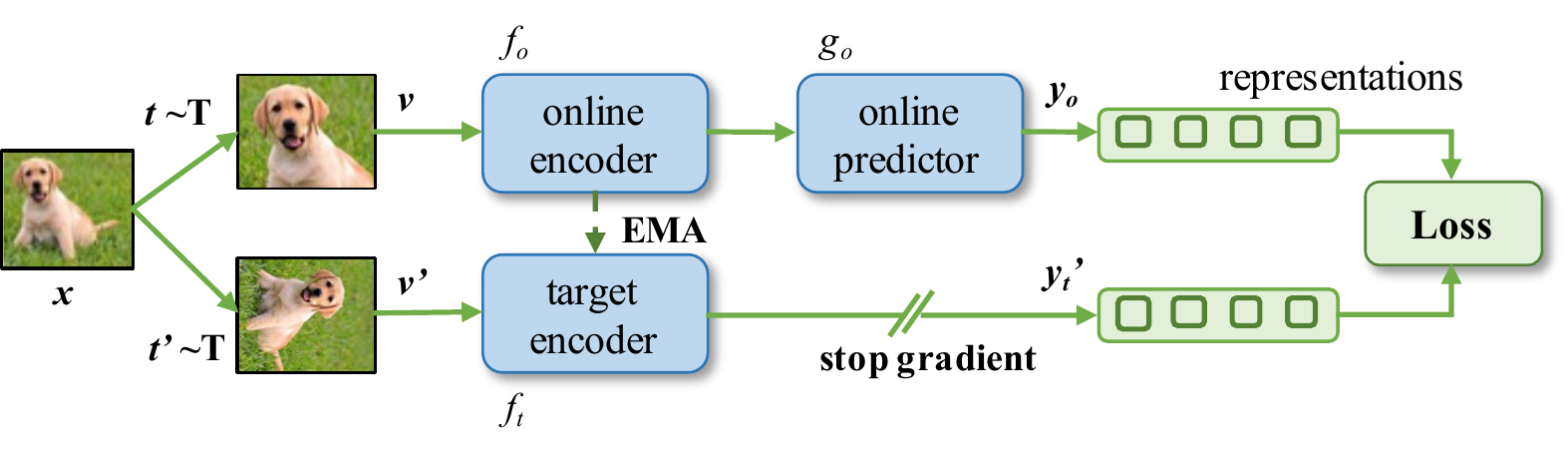}
	\caption{Illustration of contrastive learning. It comprises an online network and a target network. 
    An input image $x$ is transformed by two augmentation methods $t \sim T$ and $t' \sim T$ to produce two views. These two views are then sent into the network to generate representations $y_o$ and $y_t$. Contrastive learning minimizes a dissimilarity loss between $y_o$ and $y_t$.}
	\label{fig:contrastive}
\end{figure}

\subsection{Contrastive Learning}

Contrastive learning is a self-supervised approach to extracting visual representations from unlabeled data. It trains an encoder to provide generalized information for downstream tasks by minimizing a contrastive loss. In this work, we employ the contrastive learning approach from \cite{NIPS2020BYOL}, since it performs best when learning representations from decentralized unlabeled data collaboratively \cite{ICLR2022FedEMA}. 
In the training process, the contrastive loss is evaluated on a pair of features extracted from augmented views of one image. By optimizing the contrastive loss, contrastive learning maximizes the agreement of representations between these two views.
As shown in Fig. \ref{fig:contrastive}, contrastive learning utilizes an asymmetric network containing an online network with an online encoder $f_o$ and a predictor $g_o$ and a target network with a target encoder $f_t$. 
Optimizing the contrastive loss only contributes to the online network update, while stop gradient prevents the gradient optimization of the target network. The target network is updated by the exponential moving average (EMA) of the online network.
For an unlabeled input image $x$, two data augmentations methods $t \sim T$ and $t' \sim T$ are applied to $x$ to produce two augmented views $v = t(x)$, $v'=t'(x)$, where $T$ is a distribution of image augmentations. They are fed into the online network and target network respectively and output representations $y_o = g_o(f_o(v))$, $y_t' = f_t(v')$. The contrastive loss is the dissimilarity between the representations:
\begin{equation} \label{equ:BYOLloss}
\begin{split}
\mathcal{L}(x)=\parallel\bar{y_o}-\bar{y}_t'\parallel_2^2 = 2-2\cdot \frac{ \left \langle y_o, y_t' \right \rangle} 
  { \parallel y_o\parallel _2 \cdot \parallel y_t'\parallel _2 } \\
\end{split},
\end{equation}
where $\bar {y_o} = y_o / {\parallel y_o \parallel} _2 $ and $\bar y_t' = y_t' / {\parallel y_t' \parallel} _2 $ are the $\ell _2$-normalized representations.
This contrastive learning approach is compatible with federated learning since it doesn’t require negative pairs formed by samples in the whole training batch that is distributed on devices.

\begin{figure*}[!htb]
	\centering
	\includegraphics[width=1.0\linewidth]{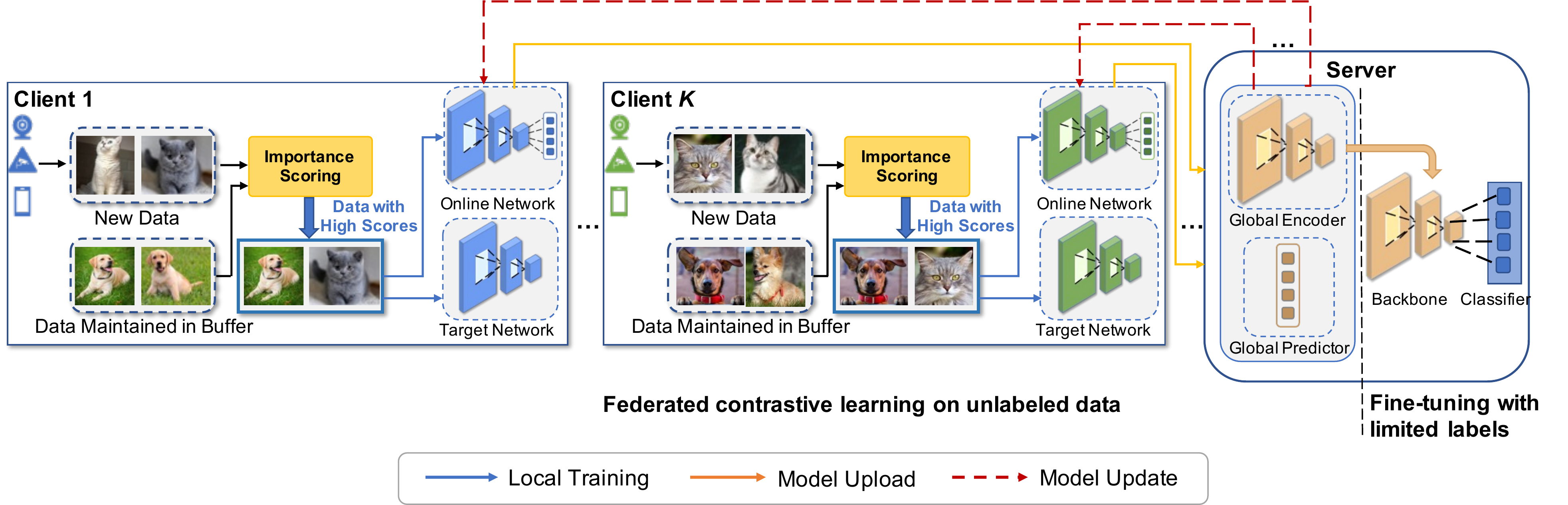}
	\caption{Overview of the proposed federated contrastive learning framework SOFed. The network contains a server and $K$ devices (clients). The online encoder is first collaboratively trained on clients with data selected from streaming unlabeled data to generate good representations. Then a classifier is trained with few labeled data on top of the online encoder.}
	\label{fig:framework_overview}
\end{figure*}

\subsection{Federated Learning}

Federated learning is a distributed training framework to train a global model without centralizing local raw data \cite{pmlr2017mcmahan}. FedAvg is a typical algorithm of federated learning methods. 
In each round of FedAvg, a subset of clients $C$ are activated to perform training on local datasets $\{D_i, i \in C \}$ and transmit the updated model parameters $\{W_i, i \in C \}$ to server. 
Then the server aggregates the updated models and updates the global model $W_g$ by averaging the local model parameters
$ W_g \gets { \sum_{i \in C}{ \frac {|D_i|}{ \sum_{i \in C}|D_i| } W_i   } } $. 
This communication round repeats to update the global model until convergence.
While the traditional federated learning methods \cite{pmlr2017mcmahan, arxiv2018zhao, ICLR2020FedMA, icml2020scaffold} have privacy guarantee, they assume the devices have enough storage resources and the input data stream is fully labeled.

\subsection{Related Work}

\textbf{Self-supervised Visual Representation Learning.}
Self-supervised learning tries to learn effective visual representations from unlabeled data \cite{ECCV2015Unsupervised, ECCV2016Unsupervised, chen2020simclr, he2020moco, NIPS2020BYOL,survey2021ssl}. Among those methods, contrastive learning \cite{chen2020simclr, he2020moco,cvpr2021simsiam, NIPS2020BYOL} demonstrates state-of-the-art performance. Some contrastive learning methods \cite{chen2020simclr, he2020moco} rely on negative pairs to prevent model collapse. For example, SimCLR produces the negative pairs from other samples in a large training batch. MoCo generates the negative pairs from a memory bank. Recently, some contrastive learning methods are proposed to eliminate the negative pairs by adopting stop-gradient operation to prevent the collapsing of Siamese networks \cite{cvpr2021simsiam, NIPS2020BYOL}.
All these methods require training on a large amount of accessible data. During training, minibatches are generated by randomly splitting the dataset and sent to the server. By reshuffling the training batches over epochs, the training data can maintain independent and identically distributed. However, in distributed on-device learning, the input streaming data distributed across devices are usually non-iid and it is not feasible to maintain a large dataset on edge devices. Simply applying contrastive learning on each device with a limited number of training data will degrade the performance. Therefore, it is necessary to conduct contrastive learning collaboratively to obtain good visual representation from online unlabeled data in a network of edge devices.



\textbf{Federated Contrastive Learning.}
There are several federated contrastive learning methods \cite{FedCA2020zhang, ICCV2021fedu, ICCAD2021Wu, ijcai2022wu} to collaboratively extract representations from unlabeled distributed data while preserving privacy. \cite{FedCA2020zhang, ICCAD2021Wu, ijcai2022wu } address the non-iid challenge by sharing features, hence, having the risk of privacy leakage. \cite{ICCV2021fedu } is based on BYOL and achieves state-of-the-art accuracy in downstream tasks. However, these methods neglect the restriction of storage resources. Therefore, existing methods cannot be applied in real-life scenarios where storage resources on edge devices are limited.



\textbf{Coreset Selection in Streaming and Continual Learning.}
Training on streaming data with non-IID distribution results in catastrophic forgetting. To overcome catastrophic forgetting, replay buffer technique is widely adopted in continual learning. Existing methods \cite{hayes2019memory, aljundi2019gradient, borsos2020coresets, wu2020enabling} select coresets from streaming data into the replay buffer to be included in the training. To overcome the catastrophic forgetting in networked scenario, some previous works \cite{infocom2021selection, arxiv2022selection, aaai2022selection} propose online data selection for federated learning. \cite{infocom2021selection, arxiv2022selection} select high-quality samples based on metrics such as gradient upper bound of each sample and the projection of the local gradient onto the global gradient. \cite{aaai2022selection} trains a relevant data selector (RDS) to select data and needs the server to intervene in the training of RDS based on the distribution of labels. All the existing approaches require data labels to select coreset. However, the requirement of labeling all the streaming data on edge devices is stringent. Hence an efficient approach to select coreset into the replay buffer in FCL is needed.




\begin{figure*}[!htb]
	\centering
	\includegraphics[width=1.0\linewidth]{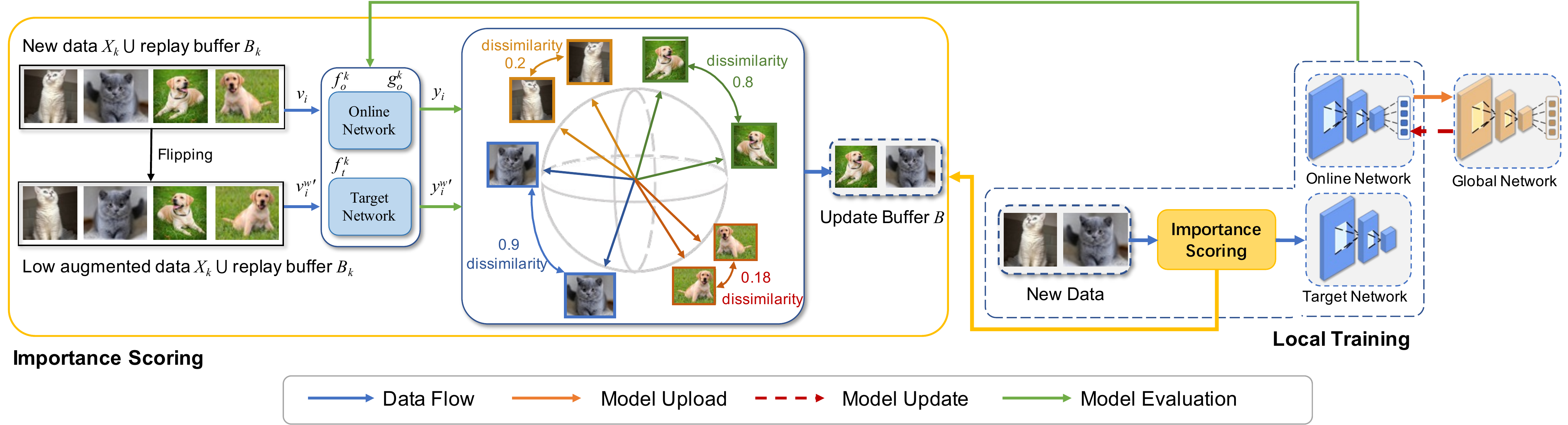}
	\caption{Local contrastive learning process with replay buffer on the $k$-th device. The replay buffer employs importance scoring for online data selection. 
    }
	\label{fig:local}
\end{figure*}

\section{Federated Self-Supervised Learning with Selective Data Contrast}\label{sec:method}

This paper proposes a framework SOFed aiming at training a global model to extract good visual representations from unlabeled streaming data in a network of edge devices with limited storage. 
This is the first work on self-supervised on-device federated learning from unlabeled streams.
To reduce the storage overhead on edge devices while learning representations efficiently, we propose a data selection method based on importance scoring. 
The data selection method facilitates the edge device to select the most representative samples into the local buffer and benefits representation learning.

In this section, we will first introduce the overview of the proposed federated contrastive learning with coreset selection framework in Section \ref{sec:overview}. Then we will present the local contrastive learning process in Section \ref{sec:local_model}. After that, we will introduce how to perform coreset selection on each device in Section \ref{sec:contrast_score}.

\subsection{Framework Overview}\label{sec:overview}


As shown in Fig. \ref{fig:framework_overview}, there are two stages in the proposed FCL framework SOFed. In the first stage, the model is collaboratively trained on streaming unlabeled data by distributed devices to generate good visual representations. An additional classifier is then trained on top of the online encoder from the learned model with few (e.g. 1\%) labeled data, in the second stage. In the rest of this paper, we focus on the FCL framework in the first stage to obtain better visual representations.

In the first stage, there is a network containing a server and many distributed devices (i.e., clients). The server is used to coordinate the process of learning, aggregate the model updates and distribute the model to clients. 
SOFed is performed round by round. 
There are four steps in one round. (1) First, the server sends the global model to devices to initialize the online network.
(2) Then, each client conducts local training.
As data keep streaming in, each device independently adopts a coreset selection policy based on importance score to maintain a replay buffer. More specifically, in round $r$, once a device $k$ receives a new batch of samples $X_k$, it evaluates the importance score of each sample in both $X_k$ and the local buffer $B_k$. For simplicity, we assume the size of $X_k$ and the size of $B_k$ are the same for each $k$. 
The device $k$ selects $|B_k|$ data with the highest scores from $B_k \cup X_k$ and updates buffer $B_k$ with these data.
With this importance-based data selection process, each device can maintain the most valuable samples from streaming data for model training, reducing storage costs and enhancing the performance of the global model simultaneously.
Local models are trained on the local replay buffers using contrastive learning to extract better visual representations.
(3) After training, the online neural networks in local models are uploaded to the server.
(4) Finally, the server aggregates the clients' online neural networks to update the global model using as in \cite{pmlr2017mcmahan}.
The details of local coreset selection and the model update will be discussed in the next subsections.


\subsection{Local Model Architecture}\label{sec:local_model}

The local contrastive learning process is shown in Fig. \ref{fig:local}. 
BYOL architecture is applied as the local model since its loss function of one sample only uses the representations of this sample and does not depend on other samples. This decoupled formulation of loss function makes it easier to measure and interpret the importance of each sample.
Different from BYOL, other popular contrastive learning models such as SimCLR, MoCo \cite{chen2020simclr, he2020moco} use other samples in the same mini-batch when computing the loss of one sample. This correlation makes it challenging to measure the importance of each sample.
To be more specific, SimCLR and MoCo use a contrastive loss which minimizes the similarity of representations between different samples:
\begin{equation} \label{equ:loss}
\begin{split}
\mathcal{L}(x)= -log \frac{ exp(sim(y_o, y_t')/\tau } 
  { \sum_{i \in Q}{ exp(sim(y_o, {y_t^i}')/\tau } } \\
\end{split},
\end{equation}
where ${y_t^{i}}'$ is the output of target network of $x_i$ in a batch or a memory bank $Q$. In contrast to (\ref{equ:BYOLloss}), the denominator in (\ref{equ:loss}) involves other samples. Therefore, this contrastive loss is prone to be affected by other samples and is not preferable to evaluate the performance of an individual sample. 

For each device $k$, the local model contains an online model with an online encoder $f_o^k$ and an online predictor $g_o^k$ and a target model with a target encoder $f_t^k$. The model parameters of the online network and the target network can be denoted as $W_o^k = (W^k, W_p^k)$ and $W_t^k$, respectively.
We utilize the online model as the main model to be aggregated at each round and as the backbone to generate representations for the second stage.
To train the online model from the input data stream, we adopt importance scoring to select the most representative data into the replay buffer and train the online model on the replay buffer.
The target model is updated by EMA with the online model: $W_t^k = \tau W_t^k+(1-\tau)W^k$, where the EMA weight $\tau$ is empirically set to 0.99.

\subsection{Local Coreset Selection Policy}\label{sec:contrast_score}

\textbf{Importance Scoring.}
We calculate the importance scores of each sample to indicate its effectiveness during training.
For each input ${x_i}$ on the ${k}$-th device, the \emph{importance scoring} function ${S(x_i)}$ aims to measure the quality of the generated representation ${{y_i} = g_{o}^{k}(f_{o}^{k}(x_i))}$. Intuitively, the input ${x_i}$ with bad representation is more valuable to be kept for further training since it will provide new knowledge for the model to learn.
To achieve this, we generate a weakly augmented view ${{x}_{i}^{w}{'}}$ of each sample ${x_i}$ from buffer ${B_k}$ and data stream ${X_k}$. 
We feed ${x_i}$ and ${x_{i}^{w}{'}}$ to the online network and the target network respectively. The output projected representations can be denoted as ${y_i}$ and ${{y}_{i}^{w}{'}}$, where ${{y}_{i}^{w}{'} = f_{t}^{k}( x_{i}^{w}{'})}$.
Ideally, if the network has learned to generate effective representations of ${x_i}$, the representations will be similar.
Hence, we use BYOL loss as the importance score of this positive pair ${\{x_{i}, {x_{i}^{w}}'\}}$.
The importance scoring function ${S(\cdot)}$ can be denoted as:
\begin{equation} \label{equ:importance_score}
\begin{split}
S(x_i)&=dissim(x_i,{x_{i}^{w}}') \\
 & =1-similarity(\bar{y_i}, \bar{y}_{i}^{w}{'}) \\
 & = 1- \frac{ \left \langle y_i, {y_i^{w}}' \right \rangle} 
  { \parallel y_i \parallel _2 \cdot \parallel {y_i^{w}}'\parallel _2 } , x_i \in \{ {B_k} \cup {X_k} \}\\
\end{split},
\end{equation}
where $\bar{y_i}$ and $\bar{y}_{i}^{w}{'}$ are $\ell_2$-normalized vectors to enforce $\parallel{y_i}\parallel _2$ = $\parallel{{y_i^{w}}'}\parallel _2$ = $1$. Therefore, the importance score $S(x_i)$ is positive and in the range [0,2]. 
The BYOL loss can be interpreted as the dissimilarity between the projected representation vectors ${y_i}$, ${{y_i^{w}}'}$ of ${x_i}$ and its weakly augmented view ${{x_i^{w}}'}$. 
If the dissimilarity between ${y_i}$ and ${{y_i^{w}}'}$ is significant, the importance score will be large. Otherwise, the importance score will be approximately zero.
From the perspective of model effectiveness, the representations generated by views of a positive pair need to be invariant to the transformation. 
Hence, the two representations generated by a more efficient model are more similar.
Since a higher importance score represents a larger dissimilarity, input ${x_i}$ with a higher score indicates that the model has not effectively learned it.
Therefore, learning from inputs with higher scores can provide more valuable information to the model.
Minimizing the contrastive loss in (\ref{equ:BYOLloss}) to update the online network with input ${x_i}$ aims at minimizing the dissimilarity between two strongly augmented views. Thus the importance score of ${x_i}$ in (\ref{equ:importance_score}) will also decrease, resulting in a lower possibility to select ${x_i}$. 
Samples that will generate good representations tend to be discarded and the more valuable samples will be kept for further learning.
In this way, the dynamics of the importance scoring also enable the online model to update with more valuable data consistently.

Note that different augmentation methods can induce variances in generating a positive pair ${\{{x}_{i}, {{x}_{i}^{w}}'\}}$ for importance scoring. However, importance scoring aims at evaluating the model effectiveness of each sample independently from transformations.
To eliminate randomness, we 
use a single augmentation method to generate weakly augmented views. Horizontal flipping is utilized as the default augmentation method.
Hence, the importance scores ${S(\cdot)}$ of different images can be compared fairly, since ${S(\cdot)}$ is only deterministic to the input image.

\textbf{Compatibility with Other Contrastive Learning Architectures.}
Although we use BYOL loss to introduce the importance score, (\ref{equ:importance_score}) can also be implemented with other contrastive learning architectures. Since contrastive learning architectures can be unified as a Siamese network containing an online network and a target network \cite{ICLR2022FedEMA}, the importance scoring in Fig. \ref{fig:local} can also be generalized to other contrastive learning frameworks. 
To be more specific, contrastive learning methods train a Siamese network constituted by an online network and a target network. For example, SimCLR \cite{chen2020simclr} method shares the same weights between these two networks, and MoCo \cite{he2020moco}, as well as BYOL, updates the target model with the moving average of the online encoder.
With different contrastive learning methods, we can feed the ${x_i}$ and its flipped view  ${{x_i^{w}}'}$ to online and target networks respectively and compute the importance score based on the projected representations as (\ref{equ:importance_score}).

\textbf{Importance Scoring Based Coreset Selection.}
The client selects the most effective samples into its buffer according to the importance scores.
Once receiving a new segment of input ${X_k}$ on device $k$, the client $k$ has to make an online decision on how to update the buffer $B_k$ with the new samples. The goal of this process is to select and store the valuable data, i.e., a coreset, into the buffer for model training in the coming rounds. To achieve this, we apply the importance scoring function $S(\cdot)$ on each sample from ${ B_k \cup X_k }$. The buffer ${B_k}$ is then updated to ${B_{k}'}$ by selecting data with the highest importance scores in ${ B_k \cup X_k }$:
\begin{equation}\label{equ:select}
{B}_{k}'=\{x_i | x_i \in {B}_{k} \cup {I}_{k}, i \in topN(\{S(x_i)\}_{i=1}^{2N}) \},
\end{equation}
where ${N}$ is the size of ${B}$ and ${X}$, and ${topN(\cdot)}$ returns the indices of $x_i$ with ${N}$ largest importance scores.

The main steps of the federated contrastive learning with importance scoring based coreset selection are summarized in Algorithm \ref{FedCoCo}. At each round, the clients conduct local training in parallel. For a client $C_k$, the local model is first initialized by the global model $(W^g, W_p^g)$. Suppose that on each client, local data have the same velocity $v$. At the $r$-th round, with a sequence of online data $X_{k,r}^i$, where $i = 1,2,...,v$, the client $C_k$ calculates the importance scores using (\ref{equ:importance_score}) of data from both $X_{k,r}^i$ and $B_k$. Then it updates the buffer $B_k$ using (\ref{equ:select}) to maintain the most important data in the buffer. The local online model $W_o^k$ = $(W^k, W_p^k)$ is trained using the local buffer $B_k$. The local target model $W_t^k$ is updated by EMA with the local online model $W^k$. After local training, each client $C_k$ uploads the online model $W^k, W_p^k$ to the server. The server aggregates the local models to update the global model $ W^g, W_p^g$ as $ W^g \gets { \sum_{k = 1}^{K}{ \frac {1}{ K } W^k } } $ and
      $ W_p^g \gets { \sum_{k = 1}^{K}{ \frac {1}{ K } W_p^k } } $.
The global model can gain better representations after $R$ communication rounds. After that, the global encoder $W^g$ can be applied to downstream tasks.

\begin{algorithm}
 \caption{SOFed FCL Framework}
 \label{FedCoCo}
 \begin{algorithmic}[1]\itemindent=-2pt
  \Require clients $\mathbb{C}=\{ C_1,...,C_K \}$ with local buffers $\{ B_1,...,B_K \}$ , learning rate $\eta$, weight decay $m$, local data velocity $v$, the number of communication rounds $R$ and EMA weight $\tau$.
  \Ensure global encoder $W^g$.
  \State Initialize $W^g, W_p^g$
    \For{$each$ $round$ $r = 1,2,…,R$}
      \For{$client$ $C_k$ $\in \mathbb{C}$ $in$ $parallel$}
        \State $ W^k, W_p^k$ $\gets$ LocalTrain($W^{g}, W_{p}^{g}$)
      \EndFor
      \State $ W^g \gets { \sum_{k = 1}^{K}{ \frac {1}{ K } W^k } } $
      \State $ W_p^g \gets { \sum_{k = 1}^{K}{ \frac {1}{ K } W_p^k } } $
    \EndFor
  \State \textbf{return} {$ W^g$}
  \State \textbf{LocalTrain$(W^{g}, W_{p}^{g})$}
  \State $ W^k$ $\gets$ $ W^g$, $ W_p^k$ $\gets$ $ W_p^g$
  \For{$each$ $X_{k,r}^i, i = 1,2,…,v$}
    \For{$each$ $ x_j $ $in$ $ X_{k,r}^i \cup B_k$} 
      \State Calculate $ S(x_j)$ using (\ref{equ:importance_score})
    \EndFor
    \State Update $ B_k$ using (\ref{equ:select})
    \State $ (W^k,W_p^k)\gets W_o^k-\eta\nabla\mathcal{L}(W_o^k;B_k)-\eta m W_o^k $
    \State $ W_t^k  \gets \tau W_t^k+(1-\tau)W^k $
  \EndFor
 \end{algorithmic}  
\end{algorithm}

\subsection{Lazy Scoring}

Due to the limited computation resources on edge devices, we utilize the lazy scoring technique to minimize the computation overhead induced by the importance scoring \cite{DAC2021Wu}. By adopting lazy scoring, the client reuses the scores of samples in the buffer. It is based on two observations. 
First, the score $S(x_i)$ of the sample $x_i$ changes slightly in successive iterations. Since the importance score is only dependent on the sample itself and the Siamese model, the score changes slowly with a slowly updated model. Therefore, the client can utilize the scores computed several iterations before.
Second, when a new data batch arrives, the importance scoring-based coreset selection method maintains most of the data in the buffer and discards most of the new data. As will be demonstrated in Section \ref{sec:experiment}, over 80\% new data are directly dropped. Since we assume the size of the new data batch is the same as the buffer, over 80\% data in the buffer are preserved at each round. Therefore, reusing the scores can reduce lots of computation.

To be more specific, on the $k$-th client, 
we recompute the score of the sample in the buffer $B_k$ every $T$ iterations, where $T$ is the interval of lazy scoring.
For the sample $x_i$ on the $k$-th client, we record the number of iterations that it has stayed in the buffer as $age(x_i)$. 
When a new data batch $X_k$ arrives, for the samples in the buffer, we only update the ones that haven't been recomputed in  $T$ iterations.
For buffer $B_k$, the scores can be updated as:
\begin{equation}\label{equ:recompute}
\begin{split}
{S}_{k}^t=
\left\{
    \begin{array}{ll}
        dissim(x_i,{x_i^w}'), & \text{if} \ age(x_i)\;mod\;T\;=\;0\\
        S_{k}^{t-1}(x_i), & \text{otherwise}
    \end{array}
\right.
\end{split},
\end{equation}
where $S_{k}^{t-1}(x_i)$ is the score of $x_i$ from the last iteration. If the score of $x_i$ needs to be updated, the score is computed by (\ref{equ:importance_score}). Otherwise, the score remains the same as the last iteration. 
For example, suppose $T$ is 5 and sample $x_i$ is in the new data batch $X_k$. When $x_i$ first gets to the $k$-th client at the $t$-th iteration, its score $S_{k}^{t}(x_i)$ is computed and restored. The age of $x_i$ is initialized as 0. In the next 4 iterations, the score of $x_i$ will not be updated, while the age of $x_i$ increases. The buffer is updated according to the old score $S_{k}^{t}(x_i)$ of $x_i$ as (\ref{equ:select}). If $x_i$ maintains in the buffer for 4 iterations, at iteration $t+5$, since $age(x_i)$ $mod$ $T$ = 0, the score of $x_i$ is updated. The buffer is updated according to the new score $S_{k}^{t+5}(x_i)$ at iteration $t+5$.
The recomputing ratio of the importance scores can be reduced to about $1/T$ by utilizing lazy scoring. Therefore, the computation cost can be reduced and the scoring efficiency can be improved.





\begin{table*}[!htb]
	\centering
	\caption{Accuracy on CIFAR-10, CIFAR-100, SVHN datasets under different non-iid settings with different FCL frameworks. The strength of Temporal Correlation (STC) is the number of consecutive data in the input stream that are from the same class. A higher STC represents a more non-iid situation. }
	\label{tab:baslines}
	\resizebox{0.75\textwidth}{!}{
	\begin{tabular}{lcccccc}
		\toprule
		\multirow{2}{*}{Method} & \multicolumn{2}{c}{CIFAR-10} &  \multicolumn{2}{c}{CIFAR-100}& \multicolumn{2}{c}{SVHN} \\
    \cline{2-7}
    & STC 500 & STC 2500 & STC 500 & STC 2500 & STC 500 & STC 2500
\\  \hline
	FedU-FIFO   & 76.61 & 73.74 & 45.99 & 46.20 & 91.72 & 88.86 \\
	FedSimCLR-FIFO & 67.21 & 65.45 & 38.37 & 38.40 & 82.73 & 81.14   \\ 
        FedU-RR      & 76.88 & 73.63 & 46.19 & 46.88 & 91.51 & 89.32 \\
	FedSimCLR-RR & 68.20 & 66.45 & 40.21 & 40.01 & 82.30 & 81.06 \\ 
	FedU-IS      & 77.71 & 74.71 & 49.24 & 48.93 & 92.10 & 87.41 \\
	FedSimCLR-IS & 73.39 & 70.50 & 44.70 & 45.26 & 88.94 & 84.51 \\ 
	\textbf{SOFed} (ours)      & \textbf{78.66} & \textbf{74.98} &          \textbf{49.58} & \textbf{49.95} 
        & \textbf{92.20} & \textbf{89.46}  \\
\hline
    BYOL (Centralized upper-bound)  & 80.50 & 76.80 & 50.45 & 50.78 & 92.61 & 89.80           \\

		\bottomrule
	\end{tabular}}
\end{table*}

\section{Experiments}\label{sec:experiment}

In this section, we evaluate the performance of the visual representation learned by SOFed on CIFAR-10, CIFAR-100, and SVHN. We first explain the experiment setup. 
Then, we evaluate the enhanced performance of the proposed FCL framework.
After that, we evaluate the improved accuracy and the higher learning speed 
of the proposed coreset selection policy in the proposed framework.
Finally, we evaluate the impacts of lazy scoring.

\subsection{Experimental Setup}

\textbf{Datasets and Evaluation Protocols.}
We use CIFAR-10, CIFAR-100 \cite{krizhevsky2009learning} and SVHN \cite{netzer2011reading} datasets to evaluate the proposed approach. CIFAR-10 and CIFAR-100 contain 10 classes and 100 classes of an equal number of images per class, respectively. SVHN consists of 73,257 training and 26,032 testing images in 10 classes.
We use the proposed SOFed method to train the model by distributed devices without labels. To evaluate the performance of the representations generated by the encoder in the trained model, we use linear evaluation as described in 
\cite{kolesnikov2019revisiting}.
We freeze the parameters in the encoder and train a classifier on top of it with 1\%, 10\%, or 100\% labeled data. We report the classification accuracy on different datasets.

\textbf{Federated and Continual Learning Setting.}
To simulate the temporally correlated data stream, the data stream is temporally ordered by class. We use Strength of Temporal Correlation (STC) to measure the level of temporal correlation of the input stream \cite{hayes2019memory}. 
STC represents the number of consecutive data in the input stream that are from the same class. A higher STC implies a higher level of temporal correlation.
We use 5 devices for federated learning and divide the training dataset into 5 partitions.
The temporally reordered dataset is randomly split and distributed to different devices. At each communication round, we set the local data velocity to be $v = \frac{\#\text{training samples}}{5}$.
To simulate a more non-IID situation, we increase the STC and distribute 2 classes and 20 classes to each client in CIFAR-10 and CIFAR-100 respectively.

\textbf{Training Setting.} \label{sec:app_training_setting}
We use ResNet-18 as the architecture for encoders and use a multi-layer perceptron (MLP) as the predictor. 
We use re{an SGD optimizer} to train the network with contrastive loss by default. For a fair comparison with other methods, we train the network for 300 rounds from scratch.
Unless otherwise specified, the learning rate $\it \eta$ is set to 0.06, and the default batch size $|B|$ is 128 with a weight decay $m$ of 0.0001. 

\textbf{Baselines.}
The proposed coreset selection policy is referred to as \emph{Importance Scoring (IS)} method.
We implement three more unlabeled data selection methods to select coreset into the{the replay buffer} from the input data stream on the fly. \emph{Random Replacement (RR)} is a continual learning variant of reservoir sampling \cite{hayes2019memory}. It randomly selects data from the input batch and the buffer to update the buffer. \emph{First-In, First-Out (FIFO) Replacement} is recently used for incremental batch learning \cite{hayes2019memory}. FIFO replacement replaces the oldest samples in the buffer with new samples. Although these two methods are straightforward, they yield superior results to maintain data without labels in continual learning \cite{chaudhry2019continual}.
\emph{K-center} is a SOTA active learning method to select coreset without labels. It solves a k-center problem in the feature space to select the data with the largest impacts \cite{iclr2018active}.
We compare the IS coreset selection policy with different data selection policies in the proposed FCL framework. 
Since SOFed is the first federated contrastive learning framework with a small data buffer on each client, 
to demonstrate the efficacy of the proposed FCL framework,
we compare the performance of SOFed with baseline methods that combine FCL frameworks with different data selection methods.
We implement two more recently proposed FCL methods. \emph{FedSimCLR} is implemented by combining FedAvg with SimCLR \cite{chen2020simclr}. \emph{FedU} \cite{ICCV2021fedu} is a SOTA FCL framework that addresses the non-iid data problem.
Among the data selection methods, IS is the best policy and RR, FIFO also have promising performance. Therefore, we combine the aforementioned FCL methods with IS, RR and FIFO for re{a fair comparison}.
We use \textbf{\emph{FedU-IS}}, \textbf{\emph{FedSimCLR-IS}}, \textbf{\emph{FedU-RR}}, \textbf{\emph{FedSimCLR-RR}}, \textbf{\emph{FedU-FIFO}}, and \textbf{\emph{FedSimCLR-FIFO}} as baselines to compare with SOFed. We also compare with a centralized contrastive learning method \textbf{\emph{BYOL}} \cite{NIPS2020BYOL}, since the local model in SOFed is based on it. 
The centralized BYOL is implemented with the proposed importance scoring method to enable an online coreset selection with temporally correlated data stream as in \cite{DAC2021Wu}.
Kindly note that this is the first work on self-supervised on-device federated learning from unlabeled streams, which is extended from our previous work in \cite{DAC2021Wu} for a single edge device. 
Different from conventional self-supervised learning settings, the streaming data are temporally correlated and the on-device buffer size is limited.
To have a fair comparison with existing works, we reproduce their results using the same setup as our methods.

\subsection{Effectiveness of SOFed}

We evaluate the proposed FCL framework SOFed and other FCL frameworks on CIFAR-10, SVHN and CIFAR-100 by training an additional classifier with 100\% labeled data. We report the accuracy of the proposed method and other baselines in Table \ref{tab:baslines}. When STC is set to 500, the data are distributed to clients in a weakly non-iid manner. When STC is set to 2500, a client is assigned to data from four classes. Therefore, the data distribution is more non-iid. On CIFAR-10 in a weakly non-iid setting, the proposed approach achieves 78.66\% top-1 accuracy, only 1.84\% below the upper bound method centralized BYOL. Although the proposed approach does not significantly outperform the combination of FedU and the proposed coreset selection policy IS (i.e., FedU-IS), SOFed can achieve competitive performance without the communication module and the empirical hyperparameter selection process in FedU. The proposed framework outperforms FedSimCLR-IS, FedSimCLR-RR, FedU-RR, FedSimCLR-FIFO, FedU-FIFO by \{5.27\%, 10.46\%, 1.78\%, 11.45\%, 2.05\%\} respectively. The proposed framework also achieves the best accuracy in non-iid setting on CIFAR-10, where it outperforms FedSimCLR-IS, FedSimCLR-RR, FedU-RR, FedSimCLR-FIFO, FedU-FIFO by \{4.48\%, 8.53\%, 1.35\%, 9.43\%, 1.24\%\} respectively. The results demonstrate that the proposed approach achieves improved accuracy on various datasets and collaborative learning settings. 
Since the experiments are conducted in an online learning scenario where temporally correlated data stream in, the vanilla method dealing with the data stream is FIFO. FIFO facilitates the models to be trained directly on the new data without selection.
So the FCL scheme without core dataset selection in an online learning scenario is the conventional FCL schemes \cite{ICCV2021fedu, ICLR2022FedEMA} combined with FIFO. 
Compared with FedSimCLR-FIFO and FedU-FIFO, the proposed method SOFed exhibits better accuracy than FCL schemes without core dataset selection.

\begin{table}[!htb]
	\centering
	\caption{
        Accuracy on CIFAR-10 with different FCL frameworks and various numbers of devices.}
	\label{tab:scalability}
        \setlength\tabcolsep{18pt}
	\resizebox{1.0\columnwidth}{!}{
	\begin{tabular}{cccc}
		\toprule
		\multirow{2}{*}{Method} & \multicolumn{3}{c}{Number of devices} \\
        \cline{2-4}
	& 10 & 50 & 100
\\  \hline
		\multicolumn{1}{l}{FedU-FIFO} & 78.75 & 70.47 & 68.12  \\
		\multicolumn{1}{l}{FedSimCLR-FIFO} & 67.43 & 61.58 & 57.39 \\ 
		\multicolumn{1}{l}{FedU-RR} & 77.25 & 75.09 & 74.53 \\
            \multicolumn{1}{l}{FedSimCLR-RR} & 69.72 & 62.95 & 58.90 \\
            \multicolumn{1}{l}{FedU-IS} & 80.26 & 78.02 & 75.76 \\
            \multicolumn{1}{l}{FedSimCLR-IS} & 73.30 & 62.09 & 57.49 \\
            \multicolumn{1}{l}{\textbf{SOFed (ours)}} & \textbf{80.79} & \textbf{78.49} & \textbf{76.46}         \\
		\bottomrule
	\end{tabular}}
\end{table}

To demonstrate the scalability of the proposed method, we evaluate the proposed FCL framework with 10, 50, and 100 clients on CIAFR-10 when STC is set to 200. 
 We report the accuracy of the proposed method and other baselines in Table \ref{tab:scalability}.
The performance of the proposed method is robust and outperforms the baselines with various numbers of devices.
When the number of devices is 10, SOFed outperforms FedSimCLR-IS, FedU-IS, FedSimCLR-RR, FedU-RR, FedSimCLR-FIFO, FedU-FIFO by \{7.49\%, 0.53\%, 11.07\%, 3.54\%, 13.36\%, 2.04\%\} respectively. 
As discussed before, although SOFed does not outperform FedU-IS with a clear margin, our proposed framework can achieve competitive performance without the empirical hyperparameter selection process in FedU. The decent performance of FedU-IS can also demonstrate the compatibility of our coreset selection policy with other FCL paradigms.
As shown in Table \ref{tab:scalability}, all the approaches degrade when the number of devices becomes larger. However, the degradation of the proposed approach is substantially smaller than other approaches. The accuracy of SOFed drops 4.33\% when the count of devices increases from 10 to 100, while the accuracy of FedSimCLR-IS drops 15.81\%.
The proposed framework is robust in different scales of edge devices network.

\subsection{Effectiveness of Coreset Selection Policy}
\textbf{Improved Accuracy with Different Labeling Ratios.} We compare the proposed coreset selection approaches by substituting the importance scoring for \emph{Random Replacement}, \emph{FIFO Replacement}, \emph{K-Center} in the proposed FCL framework. We first perform federated contrastive learning with unlabeled data under different iid settings (i.e., different STC values), then train a classifier on top of the fixed encoder with 1\% or 10\% labeled data.

\begin{table}[!htb]
	\centering
	\caption{Accuracy on CIFAR-10 dataset under different non-iid settings with different coreset selection methods. We train a classifier with 1\% and 10\% labeled data on top of the encoder trained with FCL.}
	\label{tab:label_ratio}
	\resizebox{1.0\columnwidth}{!}{
	\begin{tabular}{ccccc}
		\toprule
		\multirow{3}{*}{Method} & \multicolumn{4}{c}{CIFAR 10} \\
        \cline{2-5}
	& \multicolumn{2}{c}{STC 500} & \multicolumn{2}{c}{STC 2500}   \\
        \cline{2-5}
	& 1\% & 10\% & 1\% & 10\%
\\  \hline
		\multicolumn{1}{l}{Random Replacement} & 59.25(-2.74) & 70.53(-2.73) & 52.71(-2.65) & 66.76(-1.77)   \\
		\multicolumn{1}{l}{FIFO Replacement} & 57.78(-5.21) & 70.6(-2.60) & 53.66(-2.70) & 66.53(-2.00)   \\ 
		\multicolumn{1}{l}{K-Center} & 53.87(-8.12) & 63.44(-9.76) & 50.63(-5.73) & 62.56(-5.97)                  \\
\multicolumn{1}{l}{\textbf{Importance Scoring (ours)}} & \textbf{61.99} & \textbf{73.2} & \textbf{56.36} & \textbf{68.53}             \\
		\bottomrule
	\end{tabular}}
\end{table}

\begin{figure}[!htb]
	\centering
        \begin{subfigure}[b]{0.45\linewidth}
		\includegraphics[width=\linewidth]{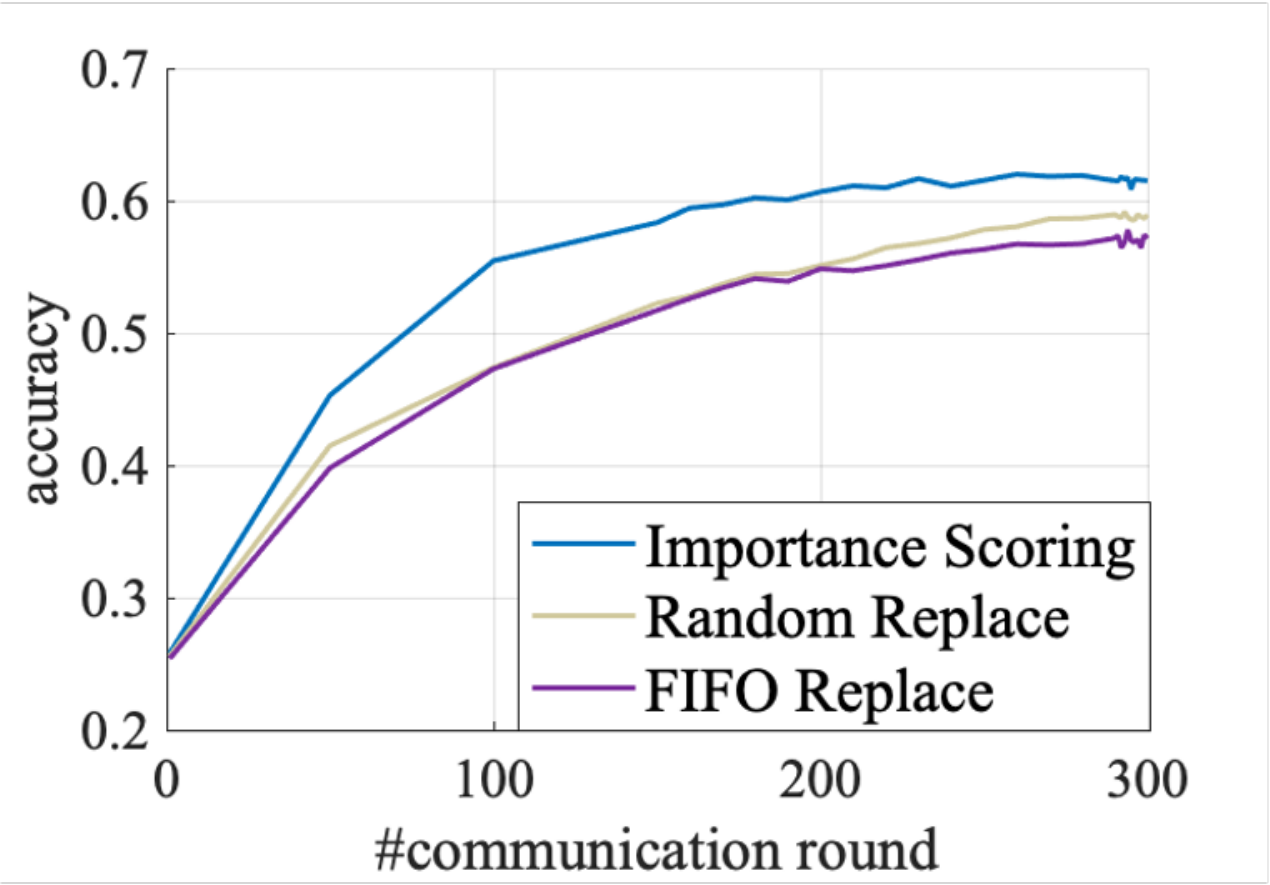}
		\vspace{-16pt}
		\caption{Accuracy with 1\% labeled data pretrained on CIFAR 10 with STC = 500.}
		\label{fig:4_1}
	\end{subfigure}
        \hspace{4pt}
	\begin{subfigure}[b]{0.45\linewidth}
		\includegraphics[width=\linewidth]{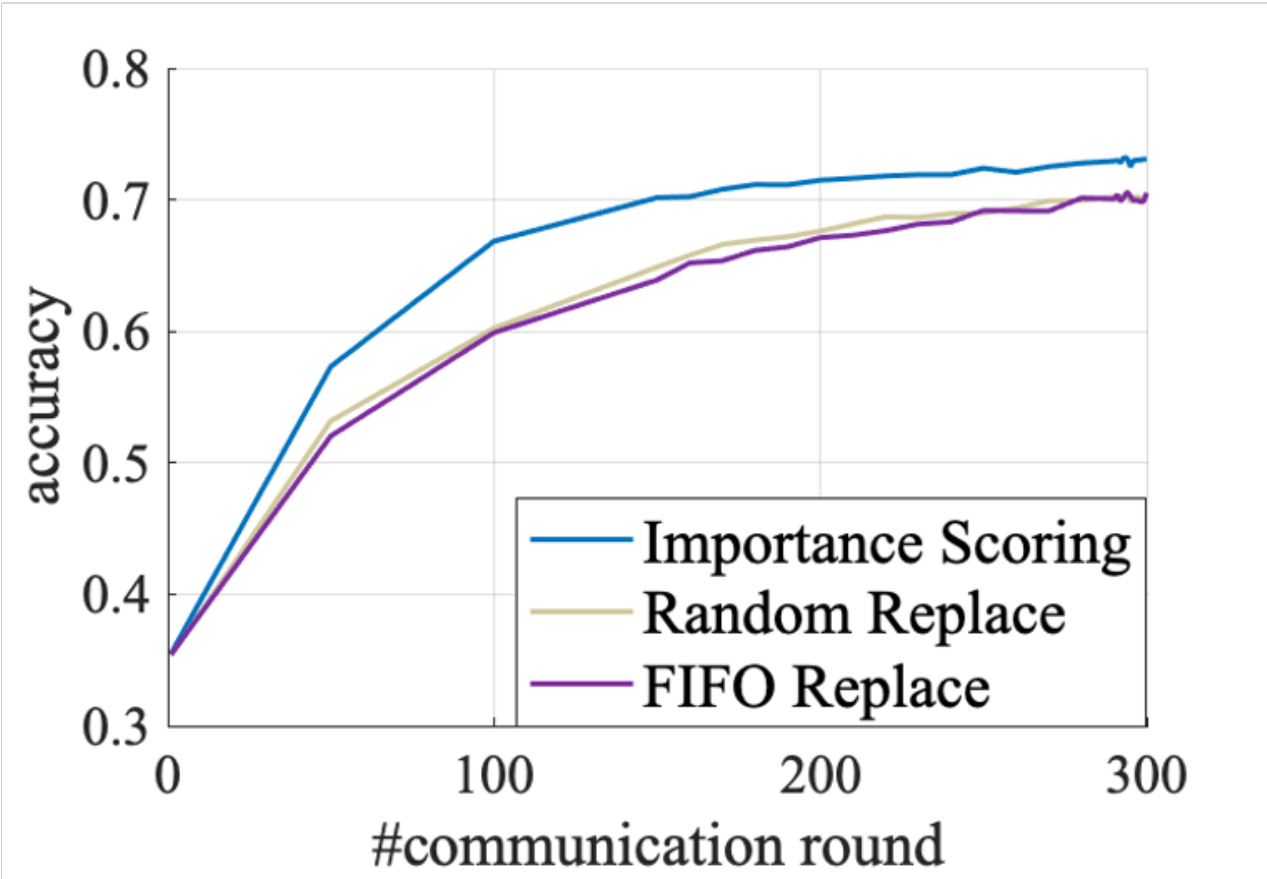}
		\vspace{-16pt}
		\caption{Accuracy with 10\% labeled data pretrained on CIFAR 10 with STC = 500.}
		\label{fig:4_2}
	\end{subfigure}
 	\begin{subfigure}[b]{0.45\linewidth}
		\includegraphics[width=\linewidth]{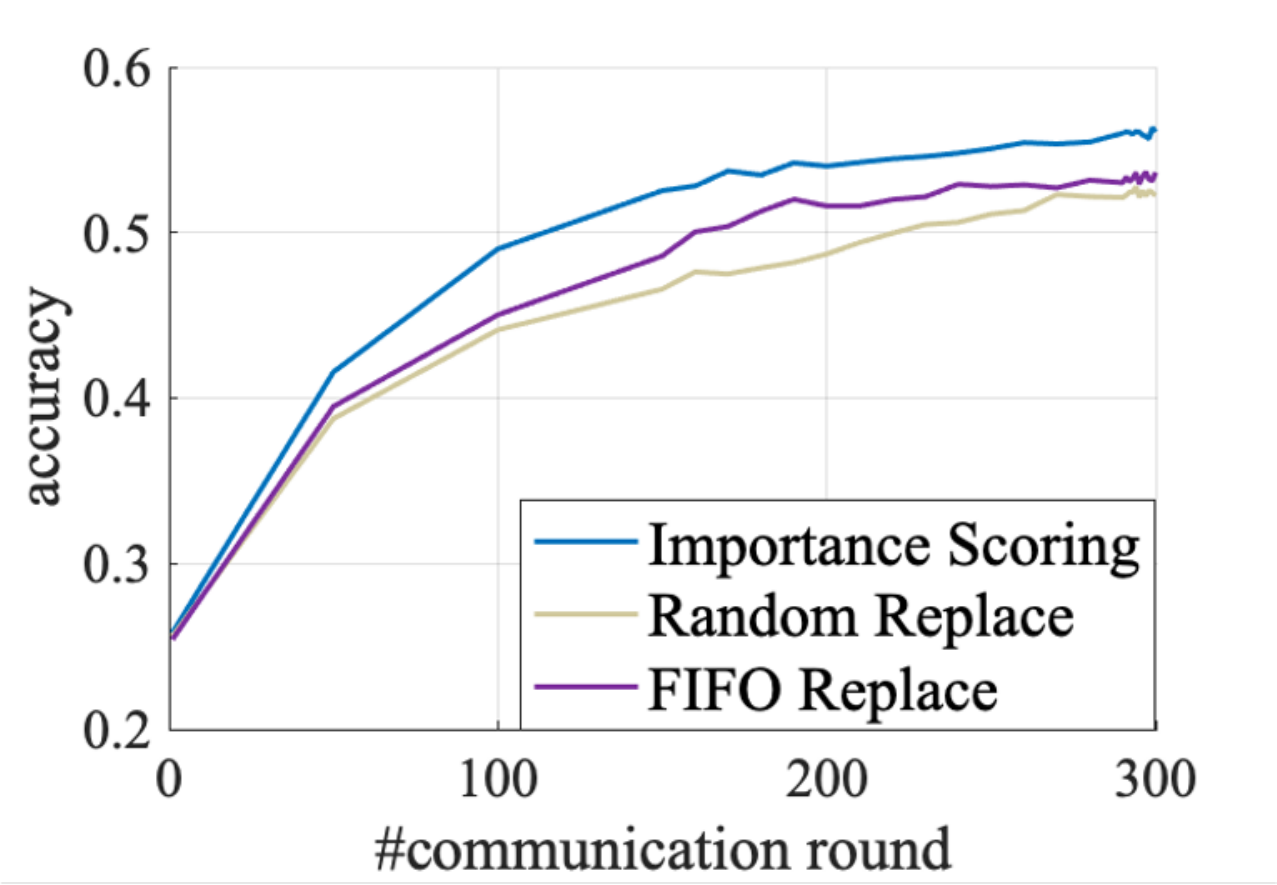}
		\vspace{-16pt}
		\caption{Accuracy with 1\% labeled data pretrained on CIFAR 10 with STC = 2500.}
		\label{fig:4_3}
	\end{subfigure}
        \hspace{4pt}
  	\begin{subfigure}[b]{0.45\linewidth}
		\includegraphics[width=\linewidth]{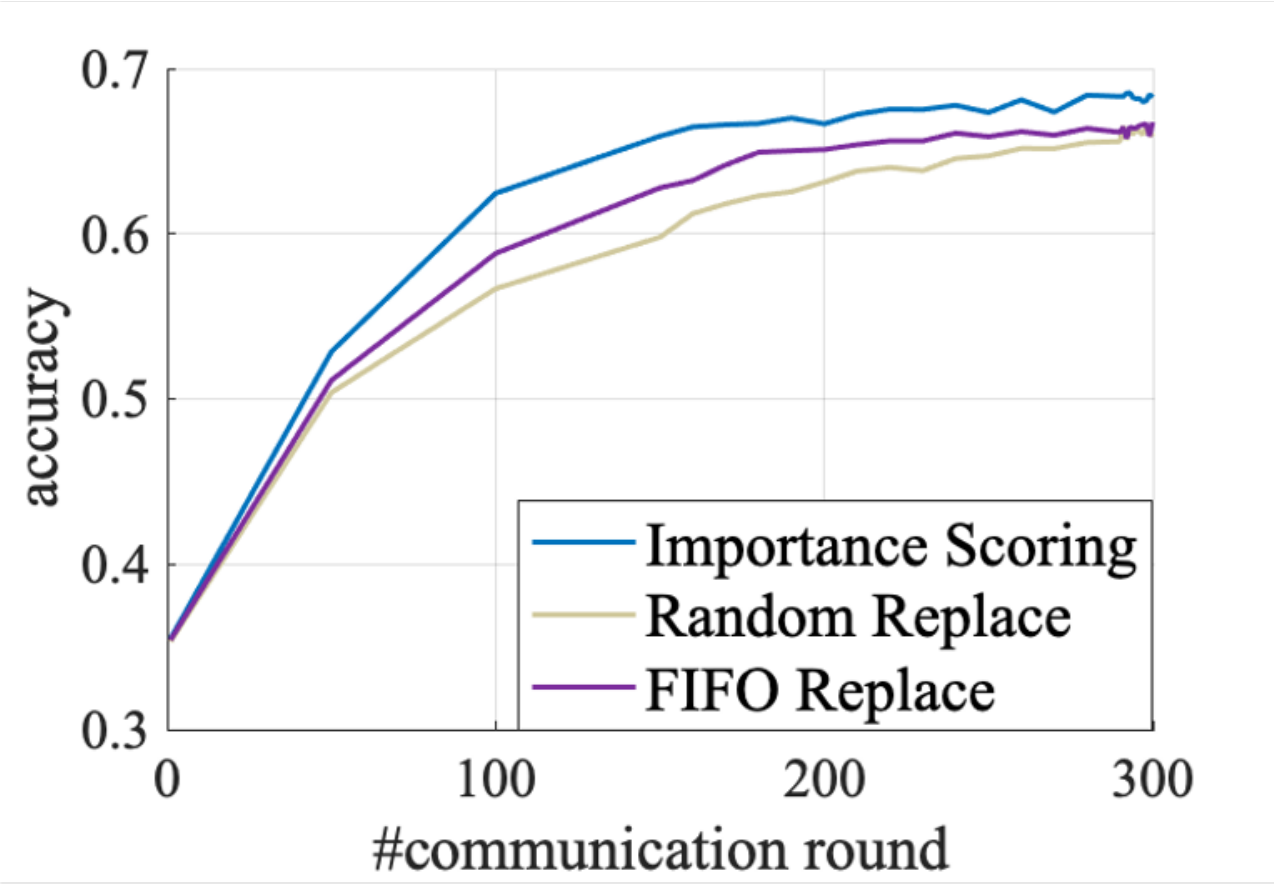}
		\vspace{-16pt}
		\caption{Accuracy with 10\% labeled data pretrained on CIFAR 10 with STC = 2500.}
		\label{fig:4_4}
	\end{subfigure}
	\caption{Accuracy on CIFAR-10 with 1\% and 10\% labeled data and different STC values.}
	\label{fig:label_ratio}
\end{figure}


The proposed coreset selection approach by importance scoring outperforms the SOTA baselines. The accuracy with different labeling ratios (i.e., 1\%, 10\%) on CIFAR-10 is shown in Table. \ref{tab:label_ratio}.
With 1\% and 10\% labeled data for learning the classifier, the proposed importance scoring achieves the accuracy of 61.99\%, 73.2\% with STC 500, 56.36\%, and 68.53\% with STC 2500. The proposed method demonstrates better accuracy in different iid settings. When STC is set to 500, importance scoring outperforms other three approaches by \{2.74\%, 5.21\%, 8.12\%\}, \{2.73\%, 2.6\%, 9.76\%\} training the classifier with 1\% or 10\% labeled data respectively. When STC is set to 2500, the proposed method outperforms other approaches by \{2.65\%, 2.7\%, 5.73\%\}, \{1.77\%, 2.0\%, 5.97\%\} with 1\% or 10\% labeled data in the linear evaluation respectively.
The proposed method learns better visual representations with different data distributions and demonstrates enhanced accuracy after training a classifier with different number of labeled data.

\begin{table}[!tb]
	\centering
	\caption{Accuracy on CIFAR-10 dataset with different buffer sizes. }
	\label{tab:buffer_size}
	\resizebox{0.8\columnwidth}{!}{
	\begin{tabular}{clc}
		\toprule
		Buffer Size          & \multicolumn{1}{c}{Method} & Accuracy     \\  \midrule
		\multirow{4}{*}{32}  
		& Random Replacement             & 67.07(-2.18)  \\
		& FIFO Replacement               & 67.72(-1.53) \\
		& K-Center                 & 62.73(-6.52)   \\ 
  &\textbf{Importance Scoring (ours) }                    & \textbf{69.25}                 \\
          \midrule
		\multirow{4}{*}{64} 
		& Random Replacement              & 67.43(-2.35)   \\
		& FIFO Replacement                & 67.63(-2.15) \\ 
		& K-Center                        & 66.65(-3.23)   \\
  & \textbf{Importance Scoring (ours) }               & \textbf{69.78}                  \\
        \midrule
		\multirow{4}{*}{128} 
		& Random Replacement             & 66.70(-3.47)   \\
		& FIFO Replacement               & 64.89(-5.28) \\ 
		& K-Center                       & 68.56(-1.61)   \\
    & \textbf{Importance Scoring (ours)}                   & \textbf{70.17} \\
        \midrule
		\multirow{4}{*}{256} 
		& Random Replacement              & 62.51(-5.69)   \\
		& FIFO Replacement                & 61.73(-6.47) \\ 
		& K-Center                        & 65.80(-2.40)   \\
  & \textbf{Importance Scoring (ours)}    & \textbf{68.20}                  \\
		\bottomrule
	\end{tabular}}
\end{table}

\begin{figure*}[!htb]
	\centering
	\vspace{-16pt}
	\begin{subfigure}[b]{0.3\linewidth}
		\includegraphics[width=\linewidth]{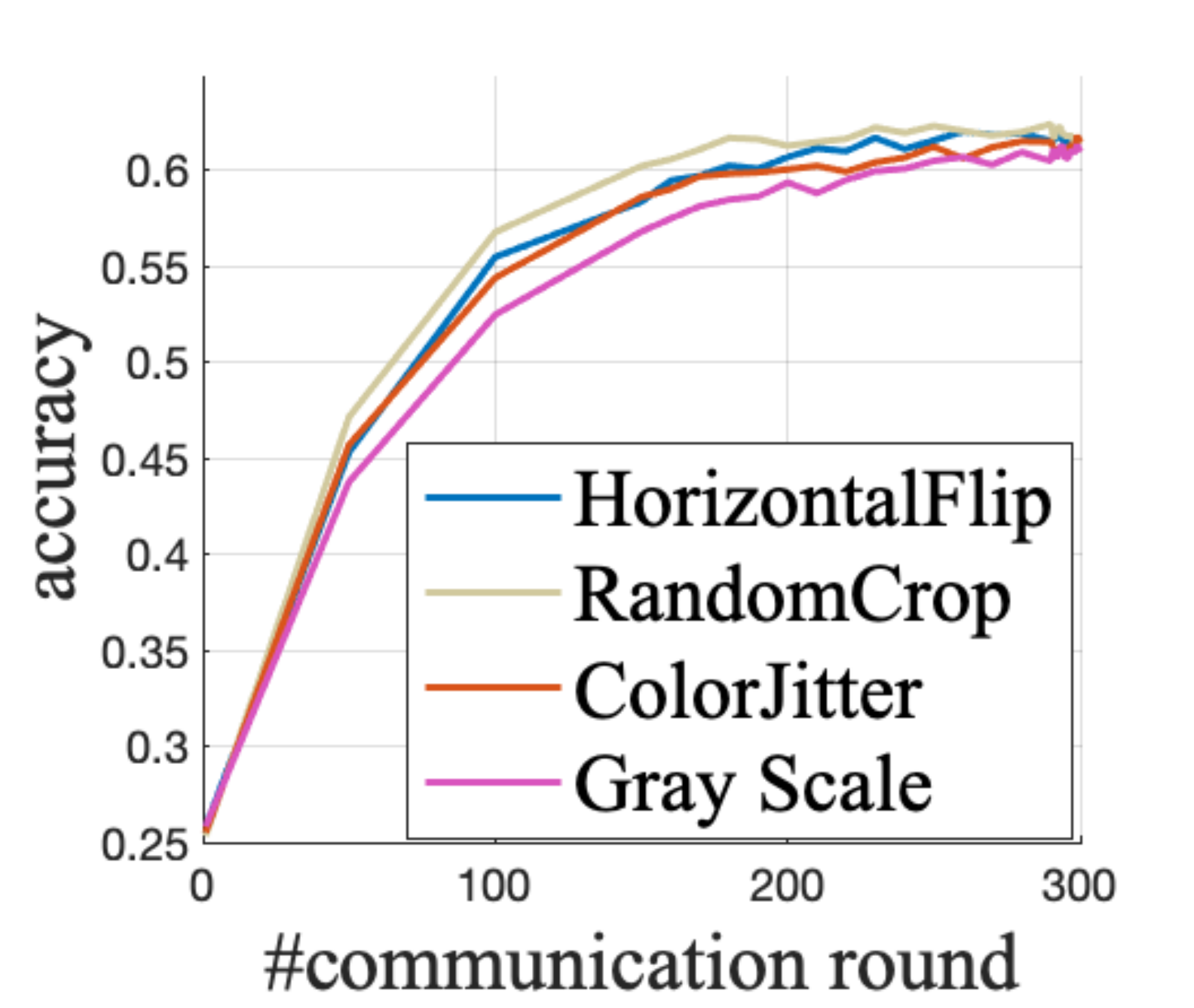}
		\vspace{-16pt}
		\caption{Accuracy with 1\% labeled data.}
		\label{fig:1}
	\end{subfigure}
	\begin{subfigure}[b]{0.3\linewidth}
		\includegraphics[width=\linewidth]{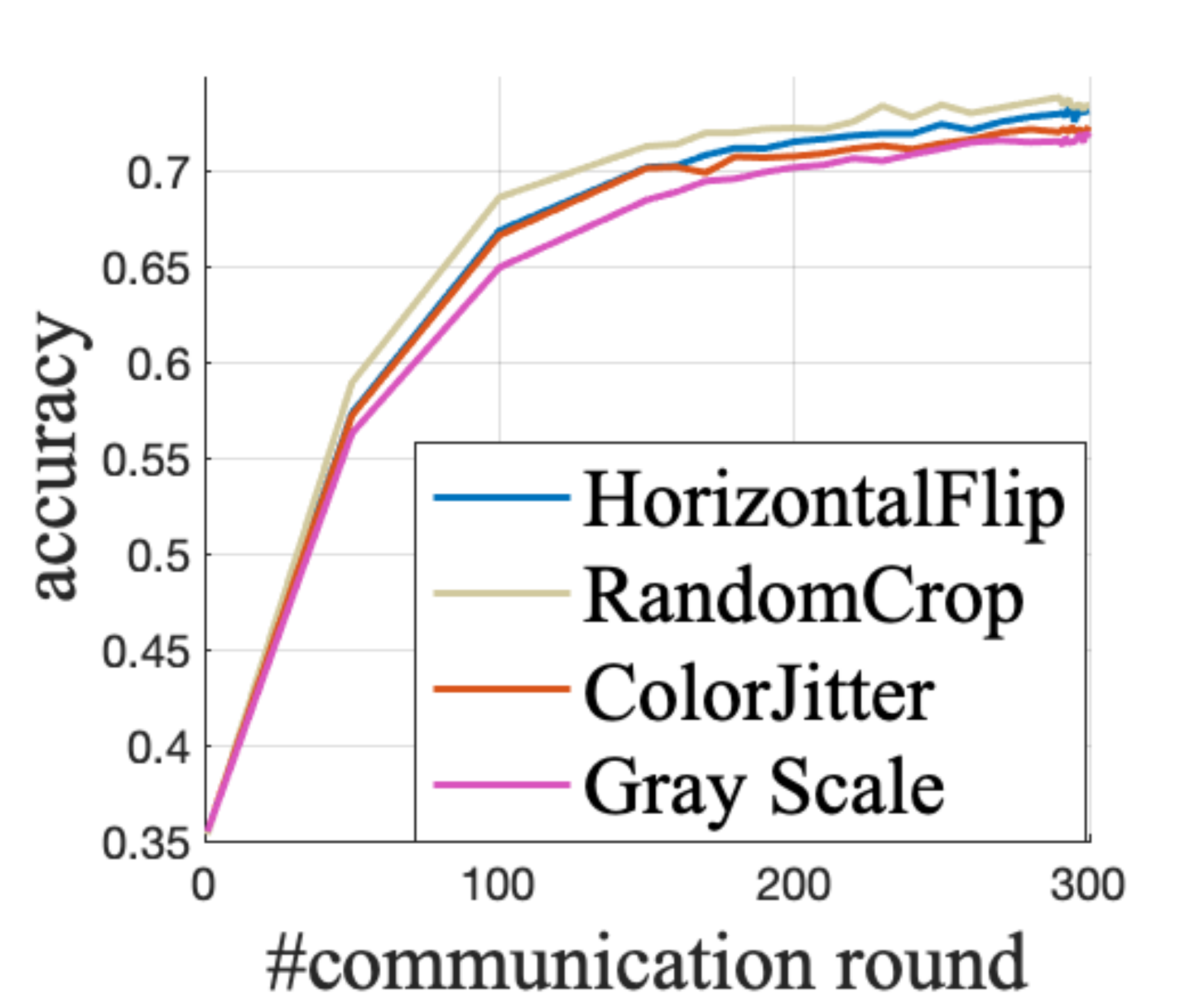}
		\vspace{-15pt}
		\caption{Accuracy with 10\% labeled data.}
		\label{fig:10}
	\end{subfigure}
 	\begin{subfigure}[b]{0.3\linewidth}
		\includegraphics[width=\linewidth]{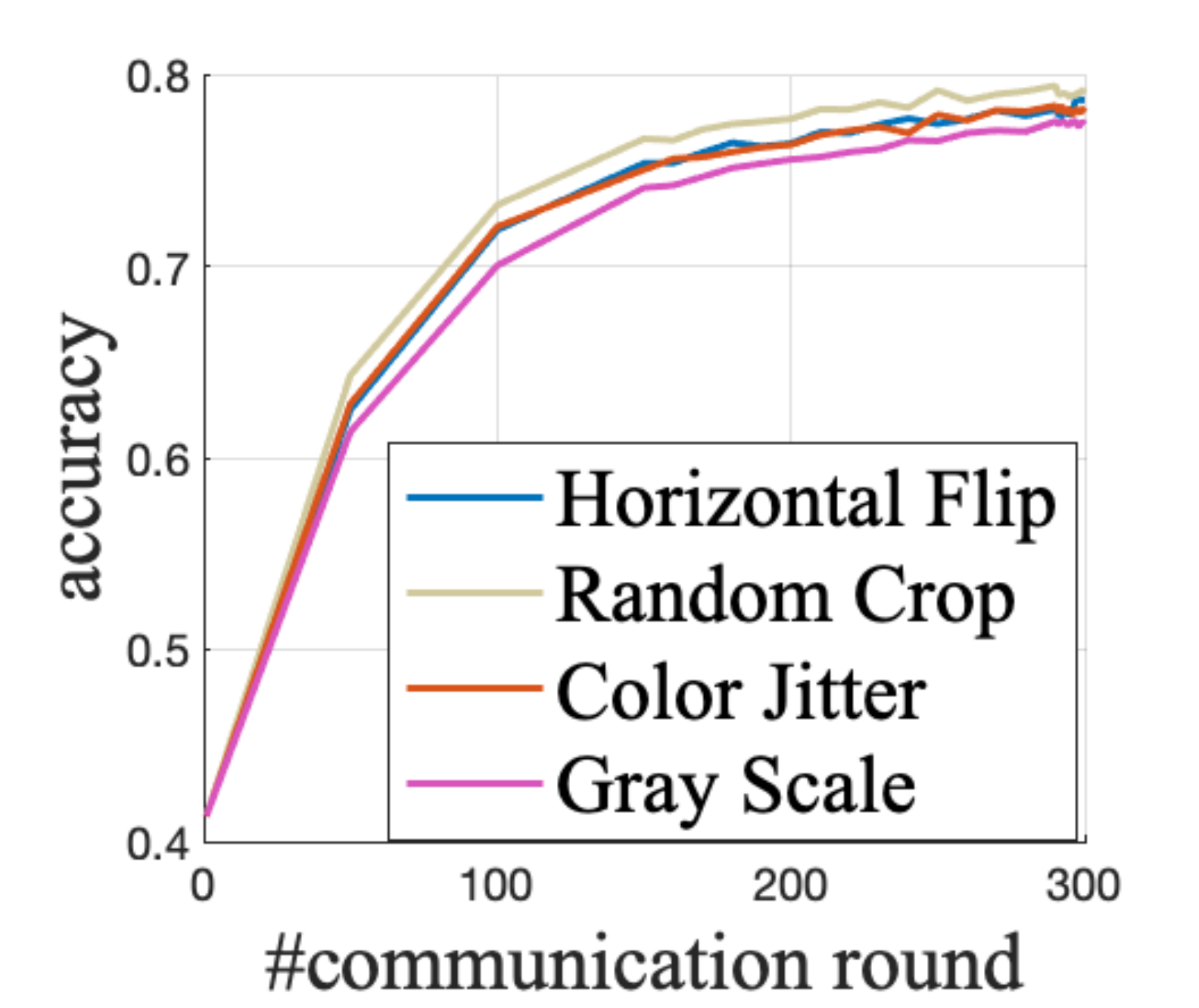}
		\vspace{-16pt}
		\caption{Accuracy with 100\% labeled data.}
            \captionsetup{labelfont={color=blue},font={color=blue}}
		\label{fig:100}
	\end{subfigure}
	\vspace{-6pt}
	\caption{Learning curves on CIFAR-10 datasets with 1\%, 10\%, and 100\% labeled data using different augmentation methods.}
	\label{fig:augmentation}
	\vspace{-8pt}
\end{figure*}

The learning curves of the proposed approach and the two best baselines trained on CIFAR-10 with different iid settings are shown in Fig. \ref{fig:label_ratio}. The accuracy is obtained by training a classifier on top of the learned visual representations with 1\% or 10\% labeled data. The learning curves represent the speed of the model to learn high-quality visual representations. We compare the proposed coreset selection method with the two most competitive baselines.
The proposed coreset selection policy quickly learns data representations and achieves a significantly faster learning speed and a higher accuracy than the baselines. When the federated contrastive learning is trained with STC equal to 500 and the linear evaluation is performed with 10\% labeled data, the accuracy of the proposed approach quickly increases to higher than 70\% at the 150-th round, which is 1.86$\times$ faster than the random replacement policy that achieves 70.12\% at the 280-th round. The proposed approach achieves a faster learning speed than the baselines.



\textbf{Improved Accuracy with Different Buffer Sizes.} We evaluate the impact of buffer size on the performance of the proposed approach. The model is trained on the CIFAR-10 dataset with re{a buffer size} in \{32, 64, 128, 256\} for 100 rounds. The corresponding learning rate is scaled to \{0.04, 0.05, 0.06, 0.08\}.

As shown in Table \ref{tab:buffer_size}, the proposed approach outperforms the baselines under different buffer sizes. The results show that with different buffer sizes, the proposed approach maintains significantly better than the baselines. When the buffer size is larger, on one hand, the framework can train the model on more informative data. On the other hand, with the same number of input data, the number of training times decreases. Hence, the accuracy doesn't change monotonically with the buffer size. The proposed method achieves the highest accuracy with a buffer size of 128.

\begin{table}[!htb]
	\centering
	\caption{Accuracy on CIFAR-10 dataset with different weak augmentation methods in importance scoring. We train a classifier with 1\%, 10\%, and 100\% labeled data on top of the encoder trained with FCL.}
	\label{tab:augmentation}
        \setlength\tabcolsep{18pt}
	\resizebox{1.0\columnwidth}{!}{
	\begin{tabular}{ccccc}
		\toprule
		\multirow{2}{*}{Augmentation Method} & \multicolumn{3}{c}{Accuracy} \\
        \cline{2-4}
	& 1\% & 10\% & 100\% 
\\  \hline
		\multicolumn{1}{l}{Horizontal Flipping} & 61.99 & 73.21 & 78.66  \\
		\multicolumn{1}{l}{Random Cropping} & 62.38 & 73.81 & 79.39  \\ 
		\multicolumn{1}{l}{Gray Scale} & 61.31 & 71.97 & 77.58 \\
\multicolumn{1}{l}{Color Jittering} & 61.79 & 72.29 & 78.39 \\
		\bottomrule
	\end{tabular}}
\end{table}

\textbf{Robustness with Different Augmentation Methods.} Since strong augmentation can introduce randomness into importance scoring, we use weak augmentation to generate views for the evaluation of importance scores. We evaluate the impact of different augmentation methods on the performance of the proposed approach. 
The model is trained on the CIFAR-10 dataset for 300 rounds with different augmentation methods in importance scoring and various percentages of labeled data in classifier training.

The selection of augmentation methods has a marginal impact on the performance of the proposed approach as shown in Table \ref{tab:augmentation}. The learning curves of the proposed framework with different augmentation methods are displayed in Fig. \ref{fig:augmentation}. When the default augmentation method, horizontal flipping, is applied as the augmentation method in importance scoring, the proposed framework achieves the accuracy of 61.99\%, 73.21\%, and 78.66\% with 1\%, 10\%, and 100\% labeled data respectively.
When random cropping is selected as the augmentation method in importance scoring, the proposed framework achieves the accuracy of 62.38\%, 73.81\%, and 79.39\% with 1\%, 10\%, and 100\% labeled data respectively.
The differences between the accuracy of these two augmentation methods are 0.39\%, 0.6\%, and 0.63\% with different percentages of labeled data.
The differences in terms of accuracy between the framework using horizontal flipping and gray scaling or color jittering are also around 1\%.
Therefore, the selection of the weak augmentation approach is insignificant in the proposed framework.



\subsection{Validation of Coreset Selection Efficacy}

\begin{figure}[!htb]
	\centering
	\includegraphics[width=0.9\linewidth]{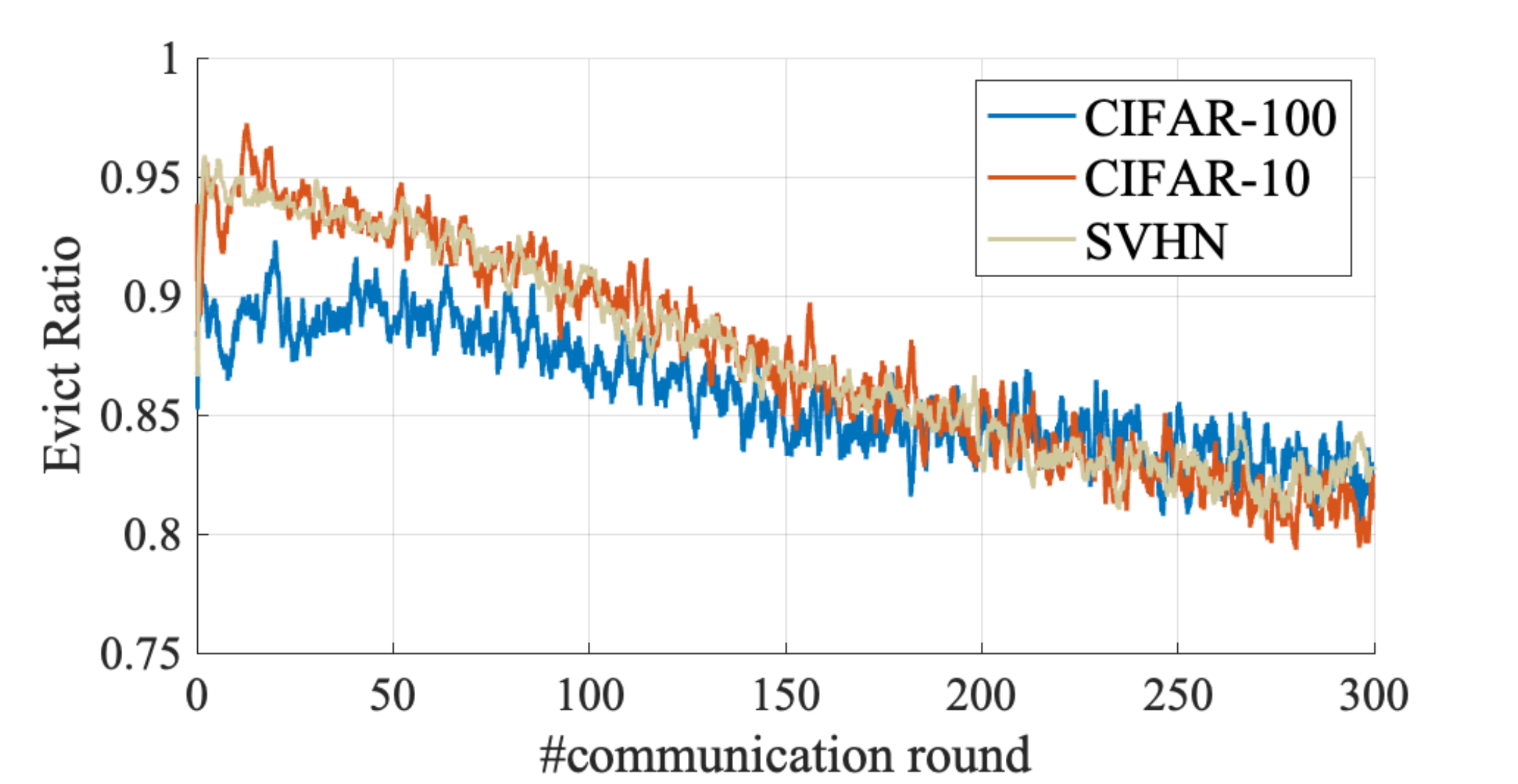}
	\caption{The ratio of new data batch directly discarded in importance scoring on CIFAR-10, CIFAR-100, and SVHN after smoothing.}
	\label{fig:evict}
\end{figure}

We evaluate the ratio of data that is directly discarded during training by using the proposed importance scoring-based coreset selection policy. The model is trained on CIFAR-10, CIFAR-100, and SVHN datasets for 300 rounds with buffer size 128 and STC value 500 in a centralized manner.
Fig. \ref{fig:evict} demonstrates the ratio of data that is directly dropped in each iteration on CIFAR-10, CIFAR-100, and SVHN datasets using importance scoring. Note that the curve of ratio with respect to iterations is smoothed using a moving average filter for better visibility. 

We can gain two insights into the coreset selection policy from these results. First, over 80\% re{of the data} in the new data batch are dropped, hence, most data in the buffer are preserved. By incorporating more information from previous iterations, the proposed coreset selection policy can preserve knowledge learned before. Therefore, the model trained on continual non-iid data stream can learn more effective visual representations. Second, the ratio of new data directly discarded decreases with respect to iteration. For Random Replacement, the ratio is around 50\% at each iteration. If the ratio is approaching 50\%, the coreset selection method could not discriminate data with various levels of importance, which implies the model has been fully trained. So the evict ratio of new data decreases during training as a result of a more accurate model.

\subsection{Impacts of Lazy Scoring}

\begin{figure}[!htb]
	\centering
	\includegraphics[width=1.0\linewidth]{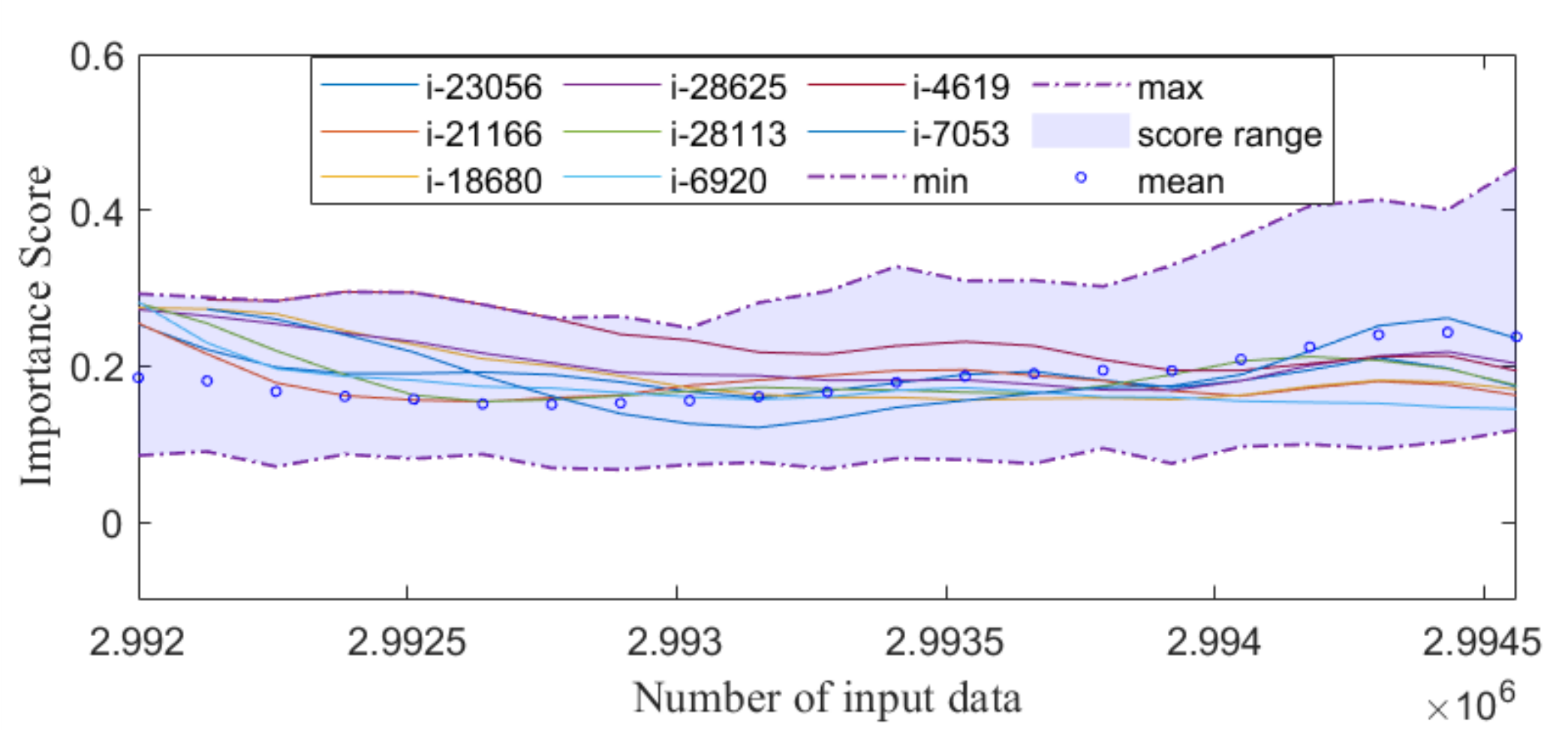}
	\caption{Importance scores of data that stay in the replay buffer longer than the lazy scoring interval. The continuous line "i-index" is the trend of importance score of the $index$-th sample. The blue shadow represents the range of scores of samples in the buffer.}
        
	\label{fig:trend}
\end{figure}

We evaluate the importance scores of the same samples in the replay buffer during training on CIFAR-10 when the lazy scoring interval $T$ is 20. 
The score of each sample changes slowly in successive iterations.
The trends of importance scores of samples in the buffer 
of the first device 
in 20 iterations are shown in Fig. \ref{fig:trend}. The $x$-axis is the number of seen inputs on one device and the $y$-axis is the importance score. 
Since some samples are discarded during these 20 iterations, for clarity, 
only the scores of the samples that stay in the replay buffer for the whole 20 iterations are depicted in the figure.
The continuous line "i-index" represents the importance score of the $index$-th sample.
The two dot lines are the minimal and maximal scores of the samples in both the buffer and the new data batch. Hence, the blue shadow represents the range of scores of samples in the buffer.
The blue circle represents the mean score of the samples in each iteration.

In the first 10 iterations, the importance scores change slowly and remain higher than the mean score. Then, the scores of the less representative samples slowly decrease. Thus, these less representative samples are recomputed and will be discarded, while some of the old and representative samples are kept in the buffer.
The results verify the motivation of the lazy scoring and rationalize its effectiveness.

\begin{table}[!htb]
	\centering
	\caption{Accuracy, rescoring ratio and batch time on CIFAR-10 dataset with different lazy scoring intervals.}
	\label{tab:lazy_scoring}
	\resizebox{0.9\columnwidth}{!}{
	\begin{tabular}{c|ccc}
		\toprule
		\multirow{2}{*}{Interval} & \multirow{2}{*}{Accuracy(\%)} & Rescoring & Relative\\
  & & Ratio(\%) & Batch Time(\%)
        \\  \midrule
		disabled  & 78.66       & 100       & 100\\
		4        & 78.48 (-0.18)       & 19.53        & 80.86 \\
            10        & 78.45 (-0.21)       & 7.42        & 74.33 \\
		20        & 79.18 (+0.52)       & 3.03        & 73.26 \\
		50    & \textbf{79.79 (+1.13)}  & 1.54         &  72.20  \\ 
            100       & 76.12 (-2.54)       & 0.73        & 70.92   \\ 
		\bottomrule
	\end{tabular}}
\end{table}

We also evaluate the accuracy and the reduced computation overhead by utilizing lazy scoring. The model is trained on the CIFAR-10 dataset with a buffer size of 128.

Table \ref{tab:lazy_scoring} demonstrates the accuracy, the rescoring ratio of buffer, and relative batch time with different lazy scoring intervals $T$. When the interval increases, the average ratio of data in the buffer that needs to be recomputed decreases. Hence, the time to process a new data batch is reduced. We aim to achieve the same level of accuracy when lazy scoring is enabled as when it is disabled. With the re-scoring interval $T$ set to smaller values 4 or 10, the accuracy is very similar to that with lazy scoring disabled, as shown in Table \ref{tab:lazy_scoring}. When $T$ is further increased to 20 and 50, lazy scoring can slightly increase accuracy. The increase in accuracy may be attributed to the incorporation of previous knowledge which could be beneficial to the data selection \cite{he2020moco}. With a lazy scoring interval of 50, the accuracy increases by 1.13\%, and the rescoring ratio is reduced to 1.54\%.
Therefore, with a proper value of $T$, the accuracy can be improved by the stability of samples in the buffer.
However, if the interval keeps increasing, the score that the sample possesses may be stale and cannot represent the sample's quality well. The accuracy decreases when the interval $T$ is set to 100, as shown in Table \ref{tab:lazy_scoring}. So the hyper-parameter $T$ should be empirically set to 50 to boost the accuracy and reduce the computation overhead at the same time.

\section{Conclusion}\label{sec:conclusion}
This work aims to enable contrastive learning from input streaming data in a network of edge devices.
We propose a framework to maintain a small data buffer filled with the most representative data for training for collaborative visual representation contrastive learning. To achieve the online selection of streaming unlabeled data, we propose a coreset selection to maintain the data buffer based on importance scores.
Experiments show superior accuracy of the proposed framework compared with state-of-the-art methods.


\bibliographystyle{IEEEtran}
\bibliography{IEEEabrv,IEEEexample}


\end{document}